\pdfoutput=1
\documentclass[a4paper]{article}

\makeatletter
\@ifundefined{Bbbk}{}{}
\makeatother

\usepackage{amsmath}
\usepackage{amssymb}
\usepackage{amsthm}

\usepackage{newtxtext,newtxmath}

\usepackage{tipa}

\usepackage[a4paper, margin=1in]{geometry}
\usepackage{graphicx}
\graphicspath{{../img/}}
\usepackage{float}
\floatstyle{plain}
\newfloat{model}{htbp}{lop}
\floatname{model}{Model}

\usepackage{booktabs}
\usepackage{multirow}
\usepackage{array}

\usepackage[caption=false]{subfig}

\usepackage{microtype}  

\usepackage{siunitx}
\usepackage[natbibapa,nodoi]{apacite}
\usepackage{hyperref}
\usepackage{placeins} 

\theoremstyle{plain}

\theoremstyle{definition}

\theoremstyle{remark}

\begin{document}


\title{Computational Typology}

\author{
Gerhard Jäger\\
University of Tübingen\\
Seminar für Sprachwissenschaft\\
Keplerstr.\ 2, 72074 Tübingen, Germany
}
\maketitle

\begin{abstract}
    Typology is a subfield of linguistics that focuses on the study and classification of languages based on their structural features. Unlike genealogical classification, which examines the historical relationships between languages, typology seeks to understand the diversity of human languages by identifying common properties and patterns, known as universals. In recent years, computational methods have played an increasingly important role in typological research, enabling the analysis of large-scale linguistic data and the testing of hypotheses about language structure and evolution. This article provides an illustration of the benefits of computational statistical modeling in typology.
\end{abstract}

\textbf{keywords:}
    typology, statistics, phylogenetics, language universals

\section{Introduction}

Typology is the subfield of linguistics studying and classifying languages according to their structural features (as opposed, e.g. according to their genealogical classification). Its aims are to delineate the diversity of human languages, and to identify common properties of all (or most) languages, so-called \emph{universals}.

The aim of \emph{statistical typology} is to identify robust correlations between typological features of languages or between linguistic and non-linguistic properties of populations and their languages, and to 
do so in a statistically sound way.

Researchers now have access to extensive datasets that enable large-scale quantitative analyses of linguistic features. These databases, such as the World Atlas of Language Structures (WALS; \citeauthor{wals13} \citeyear{wals13}), Grambank \citep{skirgardGrambankRevealsImportance2023}, Glottolog \citep{glottolog5_3}, and PHOIBLE \citep{phoible2_0}, provide detailed information on typologically relevant properties of languages. This facilitates the use of modern statistical software when conducting such analyses. In this chapter, these new developments will be reviewed and illustrated.

\subsection{Historical Context}

The origins of linguistic typology can be traced back to the early 19th century with the pioneering work of August Wilhelm von Schlegel. In 1818, Schlegel proposed a tripartite classification of languages based on morphological characteristics: fusional, agglutinative, and isolating languages. (See \citeauthor{bynon2004morphological}, \citeyear{bynon2004morphological} for details on the historical background of this classification.) Later, polysynthetic languages were added to this classification. Schlegel's work laid the groundwork for understanding how languages can be categorized based on their structural properties, such as how words are formed and how meaning is conveyed through grammar.

Schlegel's morphological types illustrated how languages could evolve from one type to another. For instance, ``a fusional language can develop into one of the isolating type, an isolating language can become agglutinative, an agglutinative language may move towards a fusional profile, and so on'' \citep[182-183]{dixon94}. Schlegel's insights highlighted the dynamic nature of language structure and set the stage for further exploration into the mechanisms of language evolution.

Georg von der Gabelentz (1840-1893) further advanced the field by emphasizing the interconnectedness of linguistic features within a language system. He argued that languages are organic systems where all parts are interdependent, and changes in one part can affect the whole. In fact, his program for linguistic typology sounds distinctly modern, as can be seen from this quotation:

\begin{quotation}
    ``But how gainful would it be if we could straightforwardly say to a language: you have this characteristic, consequently, you have those further characteristics, and that general character! – if, like the bold botanists have tried to do, we could construct the lime tree from the lime leaf. If I were allowed to baptize an unborn child, I would choose the name \emph{typology}.'' (\citeauthor{gabelentz1901}, \citeyear{gabelentz1901}, 481, translation quoted from \citeauthor{elffers2008}, \citeyear{elffers2008}, 194)
\end{quotation}

Von der Gabelentz's insights highlighted the need for a holistic approach to studying language typology, setting the stage for more systematic and statistically sound investigations. His vision of typology as a means of understanding the underlying principles of language structure was encapsulated in his metaphor of reconstructing the entire lime tree from a single leaf.

The modern era of linguistic typology began with Joseph  Greenberg's  seminal work in the mid-20th century. In his 1963 paper, ``Some Universals of Grammar with Particular Reference to the Order of Meaningful Elements,'' \cite{greenberg63} identified 45 universally or near-universally valid statements about language structure. His approach involved analyzing a diverse sample of 30 languages to uncover patterns that hold across different language families and geographical regions.

Greenberg's work introduced the concept of \emph{linguistic universals}, which can be either unconditional or conditional. Unconditional universals are features that are present in all languages, such as the tendency for subjects to precede objects in declarative sentences. Conditional universals, on the other hand, describe patterns that occur with more than chance frequency when certain conditions are met, such as the tendency for languages with Verb-Subject-Object (VSO) order to place adjectives after nouns.

\subsection{Language universals}

\cite{greenberg63} identified four types of linguistic universals based on their absolute or statistical nature and their conditional or unconditional status. These types are summarized in Table~\ref{tab:universals}.

Linguistic universals can be categorized into two main types: \emph{unconditional} and \emph{conditional} universals. Below are examples of each type, taken from \citep{greenberg63}:

\begin{table}[h!]
    \centering
    \begin{tabular}{@{}>{\raggedright\arraybackslash}p{3.5cm} >{\raggedright\arraybackslash}p{4cm} >{\raggedright\arraybackslash}p{4cm}@{}}
    \toprule
    & \textbf{Absolute (exceptionless)} & \textbf{Statistical (tendencies)} \\
    \midrule
    \textbf{Unconditional (unrestricted)} &
    Type~1. ``Unrestricted absolute universals'' \newline \textit{All languages have property X} &
    Type~2. ``Unrestricted tendencies'' \newline \textit{Most languages have property X} \\
    \addlinespace[0.5em]
    \textbf{Conditional (restricted)} &
    Type~3. ``Exceptionless implicational universals'' \newline \textit{If a language has property X, it also has property Y} &
    Type~4. ``Statistical implicational universals'' \newline \textit{If a language has property X, it will tend to have property Y} \\
    \bottomrule
    \end{tabular}
    \label{tab:universals}
    \vspace{1em}
    \caption{Typology of linguistic universals (from \citeauthor{Evans_Levinson_2009}, \citeyear{Evans_Levinson_2009}), Table 1}
\end{table}

\paragraph*{Unconditional Universals}
\begin{itemize}
    \item \textbf{Universal 1:} In declarative sentences with nominal subject and object, the dominant order is almost always one in which the subject precedes the object.
    \item \textbf{Universal 14:} In conditional statements, the conditional clause precedes the conclusion as the normal order in all languages.
    \item \textbf{Universal 35:} There is no language in which the plural does not have some nonzero allomorphs, whereas there are languages in which the singular is expressed only by zero. The dual and the trial are almost never expressed only by zero.
    \item \textit{...}
\end{itemize}

\paragraph*{Conditional Universals}
\begin{itemize}
    \item \textbf{Universal 2:} In languages with prepositions, the genitive almost always follows the governing noun, while in languages with postpositions it almost always precedes it.
    \item \textbf{Universal 3:} Languages with dominant VSO order are always prepositional.
    \item \textbf{Universal 4:} With overwhelmingly greater than chance frequency, languages with normal SOV order are postpositional.
    \item \textbf{Universal 13:} If the nominal object always precedes the verb, then verb forms subordinate to the main verb also precede it.
    \item \textit{...}
\end{itemize}

Furthermore, Greenberg distinguishes between \emph{absolute} and \emph{statistical universals}. An absolute universal applies to all languages without exception, while a statistical universal holds true for the majority of languages but may have exceptions. Here are some examples.

\paragraph*{Absolute Universals}
\begin{itemize}
    \item \textbf{Universal 14:} In conditional statements, the conditional clause precedes the conclusion as the normal order in all languages.
    \item \textbf{Universal 44:} If a language has gender distinctions in the first person, it always has gender distinctions in the second or third person, or in both.
    \item \textbf{Universal 45:} If there are any gender distinctions in the plural of the pronoun, there are some gender distinctions in the singular also.
    \item \textit{...}
\end{itemize}

\paragraph*{Statistical Universals}
\begin{itemize}
    \item \textbf{Universal 17:} With overwhelmingly more than chance frequency, languages with dominant order VSO have the adjective after the noun.
    \item \textbf{Universal 18:} When the descriptive adjective precedes the noun, the demonstrative and the numeral, with overwhelmingly more than chance frequency, do likewise.
    \item \textbf{Universal 41:} If in a language the verb follows both the nominal subject and nominal object as the dominant order, the language almost always has a case system.
    \item \textit{...}
\end{itemize}

Since Greenberg's seminal work, a plethora of similar studies with larger language samples have been conducted.

\subsection{Correlation between linguistic and non-linguistic traits}

Another major focus of research in statistical typology is the correlation between linguistic and non-linguistic traits. The idea is that languages are shaped by the social and ecological environment in which they are spoken. This is a very old idea, going back at least to the 19th century, but it has been given a new lease of life by the availability of large-scale databases that allow to test these ideas in a more systematic way.

For instance, \cite{lupyan2010language} make a strong case that languages spoken by larger populations tend to have simpler morphological structures and rely more on lexical strategies, while those spoken by smaller groups are more morphologically complex, suggesting that language structures adapt to the social environments in which they are used. \cite{atkinson2011} observes that the size of the phoneme inventory of a language is inversely correlated with the distance from Africa along the likely migration routes of early humans. \cite{hayBauer07} observe a positive correlation between population size and sound inventory size, while \cite{everettetal15} argue for a connection between climate and tonality patterns in languages.

\begin{figure}[h!]
    \centering
    \includegraphics[width=.4\textwidth]{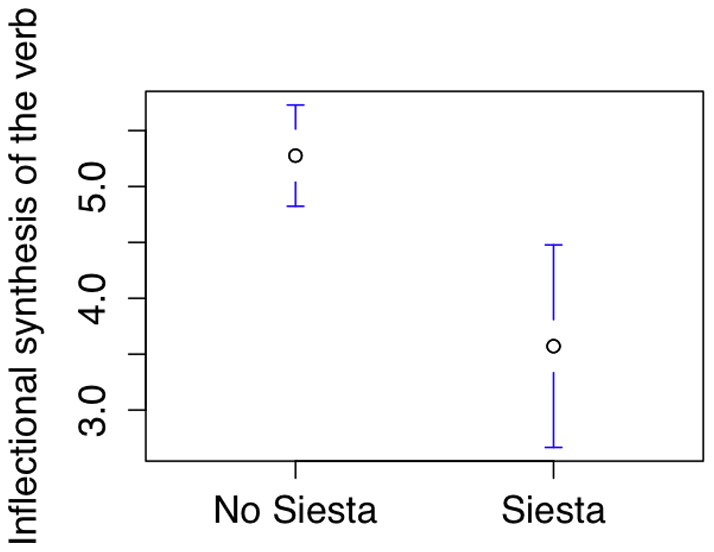}\hspace{2cm}%
    \includegraphics[width=.4\textwidth]{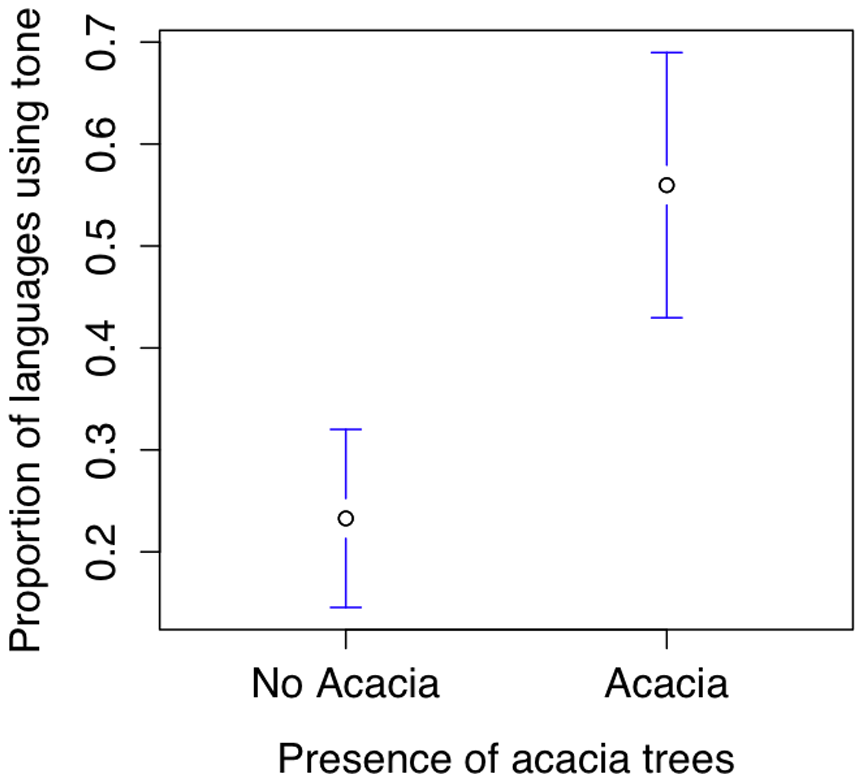}
    \caption{Correlations between linguistic and non-linguistic traits from \cite{Roberts2013}. Left: Correlation between inflectional synthesis of verbs and siesta habits. Right: Correlation between presence of tone and presence of acacia trees. Sources: doi.org/10.1371/journal.pone.0070902.g005 and doi.org/10.1371/journal.pone.0070902.g008}
    \label{fig:roberts_winters}
\end{figure}

This list can be expanded considerably. \cite{Roberts2013}, however, urge caution when digging for correlations, lest one falls into the trap of ``just-so stories'' that are not backed up by scientifically plausible causal mechanisms. To bring this point home, they list correlations like those between the inflectional synthesis of verbs in a language with the habit of its speakers to hold siestas, or the presence of tone in a language with the presence of acacia trees. The two relevant figures from their paper are reproduced in Figure \ref{fig:roberts_winters}.

\section{Electronic Resources}

The availability of large-scale databases has revolutionized the field of typology by providing researchers with access to extensive linguistic data. These databases contain information on a wide range of typological features, such as word order, case marking, and phonological systems, for hundreds of languages. Some of the most widely used databases include:

\begin{itemize}
    \item \textbf{World Atlas of Language Structures (WALS)}: This database is a key resource for understanding the structural properties of languages, including word order, case marking, and phonological systems. It is based on extensive fieldwork by typologists and is freely accessible online through the WALS website \href{https://wals.info/}{https://wals.info/}, \citep{wals13}.

    \item \textbf{Grambank}: This database focuses on the morphosyntactic properties of languages, such as word order, case marking, and agreement systems. Like WALS, it is built on fieldwork by typologists and is available online through the Grambank website \href{https://grambank.clld.org/}{https://grambank.clld.org/}, \citep{skirgardGrambankRevealsImportance2023}.

    \item \textbf{Glottolog}: Specializing in the genealogical classification of languages, Glottolog provides detailed information on language families, subfamilies, and isolates. It is based on the work of historical linguists who have reconstructed language relationships and is accessible online at \href{https://glottolog.org/}{https://glottolog.org/}, \citep{glottolog5_3}.

    \item \textbf{PHOIBLE}: This database offers comprehensive data on the phonological properties of languages, including segment inventories, syllable structures, and tone systems. It is the result of phonological analyses by experts in the field and is available online at \href{https://phoible.org/}{https://phoible.org/}, \citep{phoible2_0}.

    \item \textbf{APiCS}: Focusing on pidgin and creole languages, APiCS provides insights into their grammatical properties, such as word order, case marking, and tense-aspect systems. It is based on research by creolists and is freely available online at \href{https://apics-online.info/}{https://apics-online.info/}, \citep{apics}.

    \item \textbf{AUTOTYP}: Similar to WALS, AUTOTYP provides information on the typological properties of languages, including word order, case marking, and phonological systems. It is also based on fieldwork by typologists and is accessible online at \href{https://autotyp.uzh.ch/}{https://autotyp.uzh.ch/}, \citep{autotyp2018}.

    \item \textbf{Lexibank}: This database contains information on the lexical properties of languages, such as word lists, cognate sets, and semantic domains. It is based on the work of historical linguists who have reconstructed language relationships and is available online at \href{https://lexibank.clld.org/}{https://lexibank.clld.org/}, \citep{list2022lexibank}. Lexibank currently (March 2025) comprises about 3.5 million lexical entries from ca.\ 7,500 languages. It incorporates the data from earlier lexical data collection efforts such as the Automatic Similarity Judgment Program \citep{asjp20} and NorthEuraLex \citep{dellert2020northeuralex}.
\end{itemize}

These databases are crucial tools for linguistic research, offering a wealth of data that can be used to explore linguistic diversity and typology.

\section{Statistical non-independence}

A central methodological challenge in statistical typology is that languages do not constitute independent samples. This issue, commonly referred to as \emph{Galton's Problem} \citep[see][]{naroll1961two}, arises because similarities between languages may result from shared ancestry or contact rather than independent development. When related languages exhibit the same feature, it may have been inherited from a common ancestor. Similarly, when geographically proximate languages share features, the cause might be areal diffusion through language contact. As a result, standard statistical tests that assume independent observations are not applicable without adjustments. 

\subsection{Two case studies}

To illustrate this point, two running examples will be used -- a putative language universal, and a seeming correlation between linguistic and non-linguistic traits:

\begin{itemize}
    \item \textbf{Greenberg's Universal 27}: If a language is exclusively suffixing, it is postpositional; if it is exclusively prefixing, it is prepositional. \cite{greenberg63}
    \item \textbf{Phoneme inventory size and population size}: \cite{hayBauer07} observe a positive correlation between population size and sound inventory size.
\end{itemize}

\paragraph*{Greenberg's Universal 27}

To assess the empirical support for this statement, the values of two WALS features were extracted: feature 26 (Prefixing vs. Suffixing in Inflectional Morphology; \citeauthor{wals-26}, \citeyear{wals-26}) and feature 85 (Order of Adposition and Noun Phrase; \citeauthor{wals-85}, \citeyear{wals-85}). Languages with little or no inflectional affixation, without adpositions, or without a dominant adposition order were excluded, as were languages not covered by the phylogenetic tree used below (Section~\ref{sec:phylogenies}). This leaves 422 languages. The results are summarized in Table~\ref{tab:affix_distribution}.

\begin{table}[h!]
    \centering
    \setlength{\tabcolsep}{0.5em} 
    \renewcommand{\arraystretch}{1.2} 
        \begin{tabular}{@{}lrrr@{}}
        \toprule
        \textbf{Affix Type} & \textbf{Postposition} & \textbf{Preposition} & \textbf{\% Preposition} \\
        \midrule
        Strongly suffixing & 169 & 45 & 21 \\
        Weakly suffixing & 30 & 26 & 46 \\
        Equal prefixing and suffixing & 37 & 36 & 49 \\
        Weakly prefixing & 17 & 26 & 60 \\
        Strongly prefixing & 3 & 33 & 92 \\
        \bottomrule
        \end{tabular}
        \vspace{1em} 
    \caption{Distribution of affix types across postpositions and prepositions, with preposition percentages.}
    \label{tab:affix_distribution}
\end{table}

These numbers, strictly speaking, do neither confirm nor disconfirm Greenberg's statement, since exclusively prefixing or suffixing languages are not listed. However, the table shows something very similar in spirit: the more prefixing a language is, the more likely it is to be prepositional. The same holds for suffixing languages and postpositions.

\begin{figure}[h!]
    \centering
    \includegraphics[width=1\textwidth]{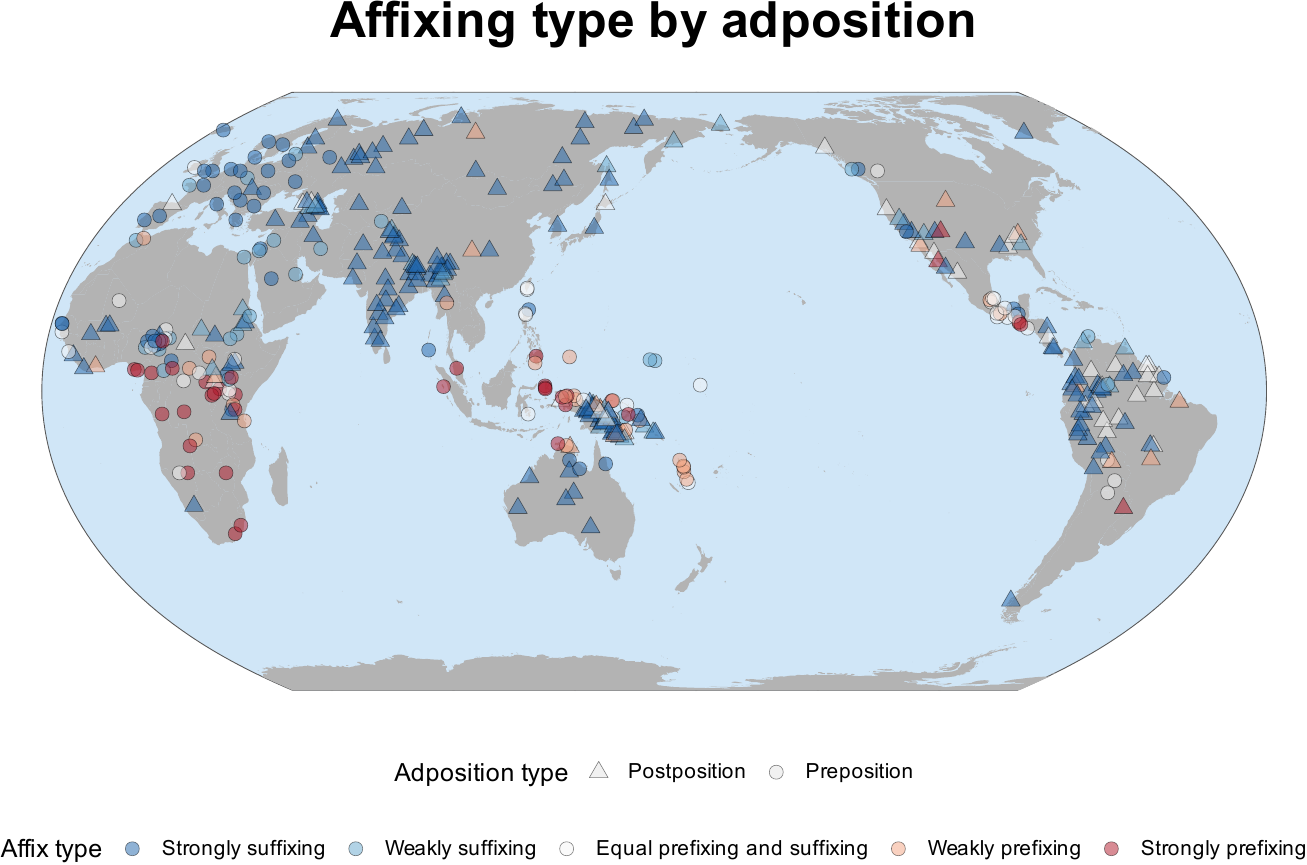}
    \caption{Distribution of affix types across postpositions and prepositions.}
    \label{fig:affix_distribution}
\end{figure}

Figure \ref{fig:affix_distribution} shows the geographic distribution of affix and adposition types.
It is clearly visible that the distribution of affix types is not random, but rather shows a clear geographical pattern. This is a clear indication that the distribution of affix types is not independent, and that this needs to be taken into account when testing for correlations.

\paragraph*{Population size and phoneme inventory size}

As mentioned above, \cite{hayBauer07} observe a positive correlation between population size and sound inventory size. To explore this effect, sound inventory size data were extracted from the PHOIBLE database \citep{phoible2_0}, and population size data were obtained from Ethnologue \citep{ethnologue16} via ASJP \citep{asjp20}.

\begin{figure}
    \centering
    \includegraphics[width=\textwidth]{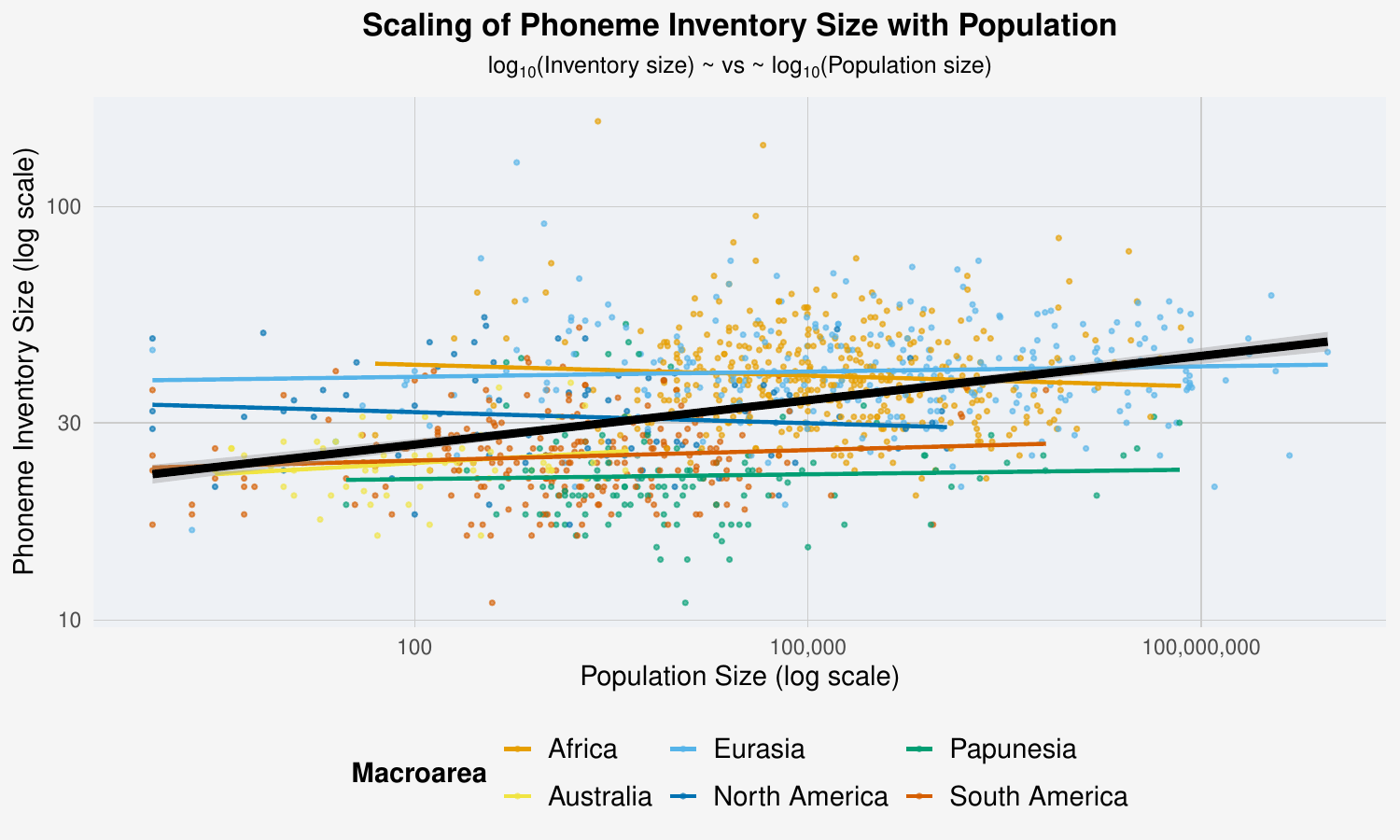}
    \caption{Scatterplot of population size and phoneme inventory size.}
    \label{fig:population_scatterplot}
\end{figure}

The scatterplot in Figure \ref{fig:population_scatterplot} shows the relationship between log-transformed population size and, also log-transformed phoneme inventory size. At first glance, there seems to be a very strong positive correlation between the two variables, as indicated by the black trend line. However, \cite{moranetal12} argue convincingly that this association is in fact a statistical artifact.

\begin{figure}
    \centering
    \includegraphics[width=\textwidth]{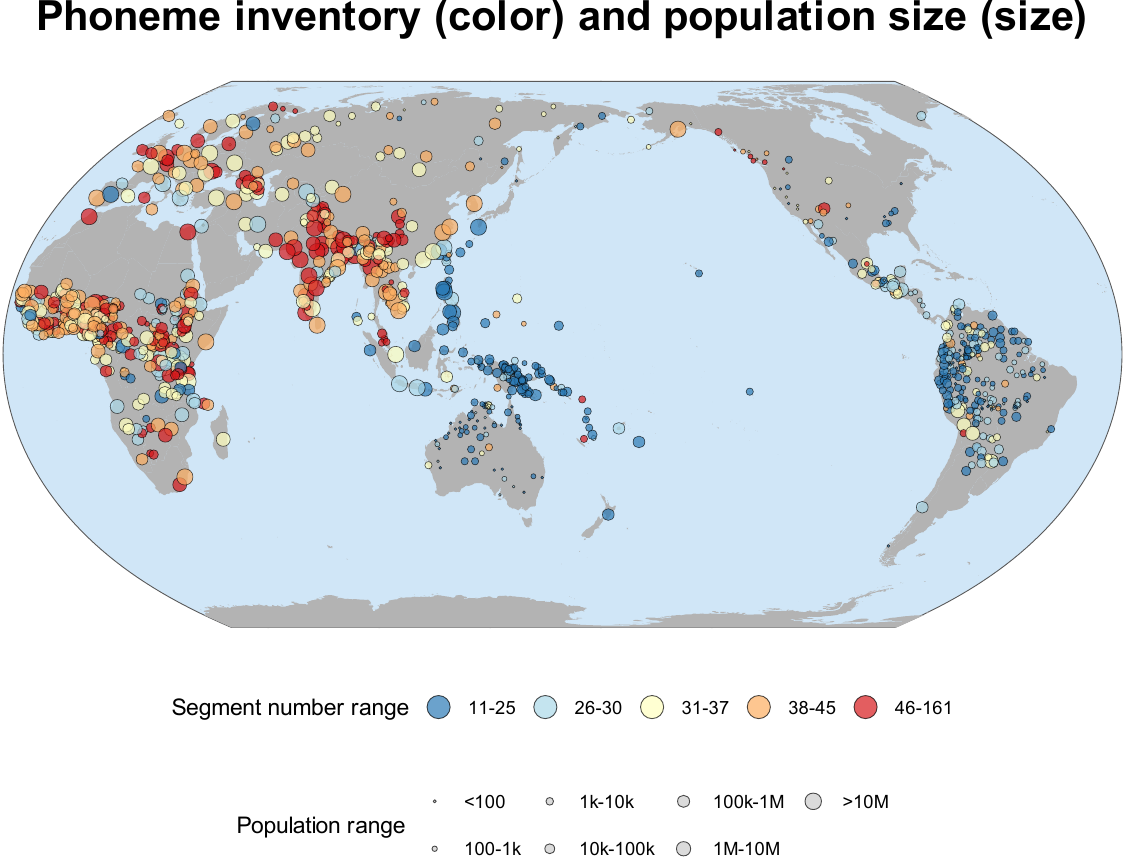}
    \caption{Map of population size and phoneme inventory size.}
    \label{fig:population_map}
\end{figure}

To get an intuition why this is so, have a look at the map in Figure \ref{fig:population_map}. It shows the population size and phoneme inventory size of the languages in the sample. The map suggests that both languages with a small sound inventory and languages with a small number of speakers are strongly concentrated in Australia/Oceania and in the Americas. 

It stands to reason that \citeauthor{hayBauer07}'s association predominantly obtains between continents rather than within continents. This impression is reinforced by the slopes of the per-macroarea trend lines shown in Figure \ref{fig:population_scatterplot}. It can be seen that there are only weak trends within macroareas, and they can be positive or negative.

This observation does not invalidate \citeauthor{hayBauer07}'s claim, but it raises the question whether the correlation is statistically significant once non-independence of languages due to geographic proximity is taken into account.

In the remainder of this section, analyses of these running examples will be compared first under the assumption of independence, and then with two more sophisticated methods taking the genealogical interdependence between languages into account. It will be demonstrated that the results are very different, and that the assumption of independence is not tenable. Similar statistical controls are possible for spatial non-independence due to contact. For reasons of space, discussion of this issue is omitted here, the interested reader is referred to \citep{GuzmnNaranjo2021StatisticalBC} for more information.

\subsection{Statistical evaluation under the assumption of independence}

As starting point, the data in the two running examples will be analyzed under the assumption that the languages in the sample are independent. This is a very strong assumption, and it is very likely that it is not true. However, it is a useful starting point to get an idea of the strength of the correlation.

All analyses were performed using the R programming language \citep{rcore2021} in combination with the statistics software \emph{Stan} \citep{stan2022} and the \emph{R}-package \emph{rstan} \cite{rstan}.

\paragraph*{Affix and adposition types}

Dependencies between two variables are often modeled via linear or logistic regression, where there is an independent and a dependent variable. The statistical model describes the influence that the independent variable has on the dependent one. This means that only the dependent variable is actually modeled; the independent variable is taken as given. While this is appropriate for experimental studies, it is not ideal for observational data such as those dealt with here. Rather, the distributions of both variables, and the dependency between them, should be modeled. This can be achieved via a \emph{bivariate} model. 

It is important to note that both variables are discrete. The affix variable is \textbf{ordinal}. Its levels are:

\begin{enumerate}
    \item Strongly suffixing
    \item Weakly suffixing
    \item Equal prefixing and suffixing
    \item Weakly prefixing
    \item Strongly prefixing
\end{enumerate}

The values are ordered, but it is not assumed that the distance between the levels is equal.

The adposition variable is \textbf{binary}, and its levels are:
\begin{enumerate}
    \item Postposition
    \item Preposition
\end{enumerate}

For both variables, the model assumes a latent continuous variable. The latent affix variable $z_{\text{affix}}$ is linked to the affix level via an \emph{ordered logistic} link function. This means that the model additionally assumes four cutpoints $c_1 < c_2 < c_3 < c_4$. The probability $F_{i,k}$ that the affix level of language $i$ is at most $k$ is the logistic transformation of $c_k - z_{\text{affix},i}$, so the larger the latent value, the more likely the higher (more prefixing) levels become. The affix level itself is a single draw from the categorical distribution over the five levels whose probabilities are the successive differences $F_{i,1}, F_{i,2} - F_{i,1}, \ldots, 1 - F_{i,4}$; it enters the likelihood as one term, not as four. The cutpoints are ordered, but the distances between them are not assumed to be equal. Rather, the positions of the cutpoints are estimated from the data. Since the cutpoints are free, they also determine the overall distribution of affix levels; a separate intercept for $z_{\text{affix}}$ would be redundant.

The latent adposition variable $z_{\text{adposition}}$ is linked to the observed level -- postposition or preposition -- via a \emph{logistic} link function. This means that the model assumes that the probability of observing a preposition equals the logistic transformation of $\alpha + z_{\text{adposition}}$, i.e., the indicator $\text{preposition}_i$ (1 if language $i$ has prepositions, 0 if it has postpositions) is a Bernoulli draw with this probability. The intercept $\alpha$ captures the overall frequency of prepositions. Without it, the model would assume a priori that prepositions and postpositions are equally frequent.

Taken together, the model can be summarized as shown in Model \ref{mod:logistic_vanilla} ($z_i$ consists of the two latent variables $z_{\text{affix}}$ and $z_{\text{adposition}}$ for language $i$).
\begin{model}[htbp]
    \centering
    \begin{minipage}{.9\linewidth}
\begin{align}
    \rho &\sim \text{Uniform}(-1, 1) \\
    \sigma_1, \sigma_2 &\sim \text{LogNormal}(0, 1) \\
    \Sigma &= \begin{pmatrix}
        \sigma_1^2 & \rho \sigma_1 \sigma_2 \\
        \rho \sigma_1 \sigma_2 & \sigma_2^2
    \end{pmatrix} \\
    z_i &\sim \mathcal{N}(\mathbf{0}, \Sigma) \\
    c_1 &\sim \mathcal{N}(0, 2) \\
    \zeta_k &\sim \text{LogNormal}(0, 2), \quad c_{k+1} = c_k + \zeta_k + 10^{-2} \\
    \alpha &\sim \mathcal{N}(0, 2) \\
    F_{i,k} &= \frac{1}{1 + e^{-(c_k - z_{\text{affix},i})}} \quad \text{for } k = 1,\ldots,4 \\
    \text{affix}_i &\sim \text{Categorical}(F_{i,1},\ F_{i,2} - F_{i,1},\ F_{i,3} - F_{i,2},\ F_{i,4} - F_{i,3},\ 1 - F_{i,4}) \\
    \text{preposition}_i &\sim \text{Bernoulli}\left(\frac{1}{1 + e^{-(\alpha + z_{\text{adposition},i})}}\right)
\end{align}
\end{minipage}
\caption{Bivariate normal model for latent variables in the affix–adposition association.}
\label{mod:logistic_vanilla}
\end{model}
The two variables $z_{\text{affix}}$ and $z_{\text{adposition}}$ follow a bivariate normal distribution with mean $0$, standard deviations $\sigma_1$ and $\sigma_2$, and a correlation coefficient $\rho$, which are estimated from the data. The small constant $10^{-2}$ keeps adjacent cutpoints apart for numerical stability.

Since each language contributes only one observation per variable, the scales $\sigma_1$ and $\sigma_2$ are only weakly identified by the data. The correlation coefficient $\rho$, however, is identified through the cross-classification of the two variables, and it is the parameter of interest here.

The model was fitted using the \emph{rstan} package in R.
\begin{table}[ht]
    \centering
    \setlength{\tabcolsep}{0.5em} 
    \renewcommand{\arraystretch}{1.2} 
        \begin{tabular}{@{}lrrrrrrr@{}} 
        \toprule
        \textbf{Parameter} & \textbf{Mean} & \textbf{SE\_Mean} & \textbf{SD} & \textbf{2.5\%} & \textbf{97.5\%} & \textbf{n\_eff} & \textbf{Rhat} \\
        \midrule
        cutpoints$_1$ & 0.08 & 0.00 & 0.33 & -0.54 & 0.81 & 21692 & 1.00 \\
        cutpoints$_2$ & 1.72 & 0.02 & 1.05 & 0.67 & 4.57 & 2120 & 1.00 \\
        cutpoints$_3$ & 4.28 & 0.06 & 2.50 & 1.91 & 11.09 & 1958 & 1.00 \\
        cutpoints$_4$ & 6.69 & 0.09 & 3.88 & 3.06 & 17.19 & 1936 & 1.00 \\
        $\alpha$ & -1.08 & 0.01 & 0.61 & -2.73 & -0.41 & 7092 & 1.00 \\
        $\sigma_1$ & 4.46 & 0.07 & 2.96 & 1.46 & 12.41 & 1869 & 1.00 \\
        $\sigma_2$ & 3.70 & 0.03 & 2.47 & 1.25 & 10.39 & 5998 & 1.00 \\
        $\rho$ & 0.79 & 0.00 & 0.12 & \textbf{0.57} & \textbf{0.99} & 3851 & 1.00 \\
        \bottomrule
        \end{tabular}
    \vspace{1em} 
    \caption{Posterior summary statistics for the bivariate model associating affix type and adposition type.}
    \label{tab:affix_bivariate}
\end{table}
The results are shown in Table \ref{tab:affix_bivariate}. The model converged successfully, as indicated by the Rhat values close to 1. The crucial outcome is the estimated value of $\rho$ of $0.79$. The $95\%$ credible interval (shown in bold) does not include zero, indicating that there is credible positive correlation between affix type and adposition type.

\paragraph*{Population size and phoneme inventory size}

As the scatterplot in Figure \ref{fig:population_scatterplot} suggests, both population sizes and sound inventory sizes are log-normally distributed; i.e., the logarithms of these quantities are approximately normally distributed. The plot also suggests that a bivariate linear model using log-transformed population size and log-transformed phoneme inventory size as associated variables is appropriate. Following standard practice \citep{mcelreath16}, both variables were standardized to have mean $0$ and standard deviation $1$. Since both variables are continuous, no latent variables and link functions are needed.

Formally, the model is specified in Model \ref{mod:bivariate_soundpop} ($x_i$ consists of the values for standardized values for log-transformed population size and log-transformed phoneme inventory size for language $i$).
\begin{model}[htbp]
    \centering
    \begin{minipage}{.9\linewidth}
        \begin{align}
            \rho &\sim \text{Uniform}(-1, 1) \\
            \sigma_1, \sigma_2 &\sim \text{LogNormal}(0, 1) \\
            \Sigma &= \begin{pmatrix}
                \sigma_1^2 & \rho \sigma_1 \sigma_2 \\
                \rho \sigma_1 \sigma_2 & \sigma_2^2
            \end{pmatrix} \\
            \mu &\sim \mathcal{N}(\mathbf{0}, 4\mathbf{I}) \\
            x_i &\sim \mathcal{N}(\mu, \Sigma) \\
        \end{align}
    \end{minipage}
\caption{Bivariate normal model for segment inventory size/population size association.}
\label{mod:bivariate_soundpop}
\end{model}

This model was fitted using the \emph{rstan} package in R, and the results are summarized in Table~\ref{tab:population_model}. The model converged successfully (as indicated by Rhat and n$\_$eff). The crucial outcome is the fact that the $95\%$ credible interval for the correlation coefficient $\rho$ does not include zero, indicating that there is a significant positive correlation between population size and phoneme inventory size.
\begin{table}[h!]
    \centering
    \begin{tabular}{@{}lrrrrrrr@{}} 
        \toprule
        \textbf{Parameter} & \textbf{Mean} & \textbf{SE\_Mean} & \textbf{SD} & \textbf{2.5\%} & \textbf{97.5\%} & \textbf{n\_eff} & \textbf{Rhat} \\
        \midrule
        $\mu_1$ & 0.00 & 0.00 & 0.029 & -0.057 & 0.057 & 20455 & 1.00 \\
        $\mu_2$ & 0.00 & 0.00 & 0.029 & -0.057 & 0.057 & 20573 & 1.00 \\
        $\sigma_1$ & 1.00 & 0.00 & 0.021 & 0.962 & 1.043 & 21378 & 1.00 \\
        $\sigma_2$ & 1.00 & 0.00 & 0.021 & 0.962 & 1.042 & 21594 & 1.00 \\
        $\rho$ & 0.39 & 0.00 & 0.025 & \textbf{0.341} & \textbf{0.437} & 19679 & 1.00 \\
        \bottomrule
    \end{tabular}
    \vspace{1em}
    \caption{Posterior summaries for the Bayesian linear bivariate model modelling log-transformed phoneme inventory size and log-transformed population size.}
    \label{tab:population_model}
\end{table}

\subsection{Taking phylogenetic non-independence into account I: Hierarchical models}

The models presented above assume that the languages in the sample are independent. However, this is a very strong assumption, and it is very likely that it is not true. In this section, it will be shown how to take this non-independence into account.

A fairly straight-foward way to do this is to use a hierarchical model, where the languages are grouped into families. (This approach has been pioneered by \citeauthor{atkinson2011}, \citeyear{atkinson2011}; see also the discussion in \citeauthor{jaegeretal2011}, \citeyear{jaegeretal2011}.) 

Each language belongs to a language family. For this study, the classification from Glottolog \citep{glottolog5_3} is assumed. Isolate languages are treated as their own family. 

Starting with the running example of the affix-adposition association, the basic setup is as above. For both features, a continuous latent variable is assumed, which is linked to the observed feature value via a link function. 

In the hierarchical model, these latent variables are composed of two components: a family-level component and a language-level component. For both components, bivariate normal distributions are assumed, with correlation coefficients $\rho_f$ and $\rho_l$ respectively. The family-level component is assumed to be the same for all languages in a family, while the language-level component is assumed to be independent for each language. 

The formal specification of the model given in Model \ref{mod:logistic_family}.

\begin{model}[htbp]
    \centering
    \begin{minipage}{.9\linewidth}
\begin{align}
    \rho_f, \rho_l &\sim \text{Uniform}(-1, 1) \\
    \sigma_{f,1}, \sigma_{f,2}, \sigma_{l,1}, \sigma_{l,2} &\sim \text{LogNormal}(0, 1) \\
    \Sigma_f &= \begin{pmatrix}
        \sigma_{f,1}^2 & \rho_f \sigma_{f,1} \sigma_{f,2} \\
        \rho_f \sigma_{f,1} \sigma_{f,2} & \sigma_{f,2}^2
    \end{pmatrix} \\
    \Sigma_l &= \begin{pmatrix}
        \sigma_{l,1}^2 & \rho_l \sigma_{l,1} \sigma_{l,2} \\
        \rho_l \sigma_{l,1} \sigma_{l,2} & \sigma_{l,2}^2
    \end{pmatrix} \\
    z_{f,j} &\sim \mathcal{N}(\mathbf{0}, \Sigma_f) \quad \text{for each family } j \\
    z_{l,i} &\sim \mathcal{N}(\mathbf{0}, \Sigma_l) \\
    z_i & := z_{f,f_{i}} + z_{l,i} \\
    c_1 &\sim \mathcal{N}(0, 2) \\
    \zeta_k &\sim \text{LogNormal}(0, 2), \quad c_{k+1} = c_k + \zeta_k + 10^{-2} \\
    \alpha &\sim \mathcal{N}(0, 2) \\
    F_{i,k} &= \frac{1}{1 + e^{-(c_k - z_{\text{affix},i})}} \quad \text{for } k = 1,\ldots,4 \\
    \text{affix}_i &\sim \text{Categorical}(F_{i,1},\ F_{i,2} - F_{i,1},\ F_{i,3} - F_{i,2},\ F_{i,4} - F_{i,3},\ 1 - F_{i,4}) \\
    \text{preposition}_i &\sim \text{Bernoulli}\left(\frac{1}{1 + e^{-(\alpha + z_{\text{adposition},i})}}\right)
\end{align}
\end{minipage}
\caption{Hierarchical bivariate model with family-level random effects for affix–adposition association. $f_i$ denotes the family of language $i$.}
\label{mod:logistic_family}
\end{model}

\begin{table}[h!]
    \centering
    \begin{tabular}{@{}lrrrrrrr@{}}
        \toprule
        \textbf{Parameter} & \textbf{Mean} & \textbf{SE\_Mean} & \textbf{SD} & \textbf{2.5\%} & \textbf{97.5\%} & \textbf{n\_eff} & \textbf{Rhat} \\
        \midrule
        cutpoints$_1$ & 0.251 & 0.002 & 0.367 & -0.421 & 1.027 & 24925 & 1.00 \\
        cutpoints$_2$ & 1.579 & 0.009 & 0.588 & 0.798 & 3.035 & 4244 & 1.00 \\
        cutpoints$_3$ & 3.522 & 0.020 & 1.098 & 2.393 & 6.450 & 3012 & 1.00 \\
        cutpoints$_4$ & 5.186 & 0.030 & 1.574 & 3.713 & 9.454 & 2832 & 1.00 \\
        $\alpha$ & -2.937 & 0.004 & 0.728 & -4.588 & -1.739 & 30423 & 1.00 \\
        $\sigma_{f,1}$ & 2.929 & 0.016 & 0.920 & 1.938 & 5.346 & 3183 & 1.00 \\
        $\sigma_{f,2}$ & 4.210 & 0.006 & 0.996 & 2.711 & 6.562 & 27681 & 1.00 \\
        $\sigma_{l,1}$ & 1.100 & 0.018 & 0.937 & 0.145 & 3.618 & 2648 & 1.00 \\
        $\sigma_{l,2}$ & 0.886 & 0.004 & 0.579 & 0.143 & 2.330 & 18131 & 1.00 \\
        $\rho_{l}$ & 0.504 & 0.003 & 0.431 & \textbf{-0.671} & \textbf{0.983} & 17564 & 1.00 \\
        $\rho_{f}$ & 0.706 & 0.001 & 0.094 & \textbf{0.491} & \textbf{0.859} & 31902 & 1.00 \\
        \bottomrule
    \end{tabular}
    \vspace{1em}
    \caption{Posterior summaries for the ordinal–binary bivariate model with family-level and language-level covariance. Cutpoints are used to map a latent continuous variable to ordinal affix positions; correlations are estimated separately at the family ($f$) and language ($l$) level.}
    \label{tab:logistic_model_family}
\end{table}

This model was fitted using \emph{rstan}. The results are summarized in Table~\ref{tab:logistic_model_family}. The crucial outcome is the fact that the $95\%$ credible interval for the family-level correlation coefficient $\rho_f$, $(0.491, 0.859)$, is positive and does not include zero. The language-level correlation $\rho_l$ is only weakly constrained by the data; its posterior mean is positive ($0.504$), but its credible interval $(-0.671, 0.983)$ is very wide and includes zero. This indicates that there is a significant positive correlation between affix type and adposition type across families, while this model cannot determine whether the two features are also correlated within families.

The hierarchical model provides a massively better fit of the data, with a log-Bayes factor of approximately 170 in favor of the hierarchical model.

Likewise, a hierarchical model was fitted to the population size and phoneme inventory size data. The model specification is given in Model \ref{mod:soundpop_family}.

\begin{model}[htbp]
    \centering
    \begin{minipage}{.9\linewidth}
\begin{align}
    \rho_f, \rho_l &\sim \text{Uniform}(-1, 1) \\
    \sigma_{f,1}, \sigma_{f,2}, \sigma_{l,1}, \sigma_{l,2} &\sim \text{LogNormal}(0, 1) \\
    \Sigma_f &= \begin{pmatrix}
        \sigma_{f,1}^2 & \rho_f \sigma_{f,1} \sigma_{f,2} \\
        \rho_f \sigma_{f,1} \sigma_{f,2} & \sigma_{f,2}^2
    \end{pmatrix} \\
    \Sigma_l &= \begin{pmatrix}
        \sigma_{l,1}^2 & \rho_l \sigma_{l,1} \sigma_{l,2} \\
        \rho_l \sigma_{l,1} \sigma_{l,2} & \sigma_{l,2}^2
    \end{pmatrix} \\
    \mu &\sim \mathcal{N}(\mathbf{0}, 4\mathbf{I}) \\
    z_{f,j} &\sim \mathcal{N}(\mathbf{0}, \Sigma_f) \quad \text{for each family } j \\
    z_{l,i} &\sim \mathcal{N}(\mathbf{0}, \Sigma_l) \\
    x_i &:= \mu + z_{f,f_i} + z_{l,i} 
\end{align}
\end{minipage}
\caption{Bivariate model with family-level random intercepts for segment inventory/population size association.}
\label{mod:soundpop_family}
\end{model}
\begin{table}[h!]
    \centering
    \begin{tabular}{@{}lrrrrrrr@{}}
        \toprule
        \textbf{Parameter} & \textbf{Mean} & \textbf{SE\_Mean} & \textbf{SD} & \textbf{2.5\%} & \textbf{97.5\%} & \textbf{n\_eff} & \textbf{Rhat} \\
        \midrule
        $\mu_1$ & -0.494 & 0.001 & 0.063 & -0.619 & -0.370 & 2797 & 1.00 \\
        $\mu_2$ & -0.325 & 0.001 & 0.072 & -0.469 & -0.182 & 2506 & 1.00 \\
        $\sigma_{l,1}$ & 0.668 & 0.000 & 0.015 & 0.640 & 0.698 & 25769 & 1.00 \\
        $\sigma_{l,2}$ & 0.658 & 0.000 & 0.015 & 0.630 & 0.687 & 26397 & 1.00 \\
        $\sigma_{f,1}$ & 0.693 & 0.000 & 0.052 & 0.597 & 0.800 & 10667 & 1.00 \\
        $\sigma_{f,2}$ & 0.860 & 0.001 & 0.062 & 0.746 & 0.986 & 13196 & 1.00 \\
        $\rho_l$ & -0.068 & 0.000 & 0.031 & \textbf{-0.129} & \textbf{-0.006} & 25858 & 1.00 \\
        $\rho_f$ & 0.583 & 0.001 & 0.083 & \textbf{0.407} & \textbf{0.731} & 6049 & 1.00 \\
        \bottomrule
    \end{tabular}
    \vspace{1em}
    \caption{Posterior summaries for the extended bivariate model with both residual- and family-level covariance components.}
    \label{tab:log_segments_population}
\end{table}
The fitted model is summarized in Table~\ref{tab:log_segments_population}. 

There are several noteworthy points to be made. Unlike in the non-hierarchical model, the estimated posterior means ($\mu_1$ and $\mu_2$) are much smaller than 0. Converting back from the normalized log-scales to the original scales, they are approximately at 3,800 population size and 29 segments. This reflects the fact that languages from large families have lower weight and isolates a higher weight in the hierarchical model.

Even more noteworthy, one finds that the language-level correlation coefficient $\rho_l$ is close to zero and, if anything, slightly negative (posterior mean $-0.068$, $95\%$ credible interval $(-0.129, -0.006)$), while one finds a pronounced positive correlation at the family level, with a posterior mean of $0.583$ and a $95\%$ credible interval of $(0.407, 0.731)$. This indicates that the correlation between population size and phoneme inventory size is driven by the family-level structure of the data.

A discussion of the implications of this finding are deferred to the next subsection.

\subsection{Taking genetic non-independence into account II: Phylogenetic models}
\label{sec:phylogenies}

Family-level random intercepts capture only a part of the genetic non-independence of languages. They have two main shortcominngs. First, they ignore the internal structure of language families. For instance, the dependency between Spanish and Portuguese is modeled as equally strong as the dependency between Spanish and Greek, since both pairs are in the same family. This can be mitigated to a certain degree by adding sub-family as random effects, but this still does not capture the intricate structure of language families.

Second, family-level random intercepts assume the degree of dependency within families is the same for all families. This ignores the difference between shallow families such as Mongolic (estimated age of divergence 700-800 years; \citeauthor{janhunen2024mongolic}, \citeyear{janhunen2024mongolic}) and deep families such as Afro-Asiatic (estimated age of divergence earlier than 10,000 years; \citeauthor{ehret1979antiquity}, \citeyear{ehret1979antiquity}).

To address these limitations, researchers increasingly turn to \textbf{phylogenetic comparative methods} (PCMs; see \citealt{harmon2019phylogenetic} for a book-length overview), which explicitly model the evolutionary relationships among languages based on a phylogenetic tree. These methods treat languages (or species, or whatever empirical phenomena are being modeled) not as independent datapoints, but as part of an evolutionary process shaped by descent from common ancestors. The phylogenetic tree encodes these relationships, including both the topology (who is related to whom) and branch lengths (how much change has happened since divergence).

Constructing family trees of languages is one of the core tasks of classical historical linguistics. However, the trees produced by the comparative method are only of limited use for PCMs, since (a) they do not include branch lengths, and (b) they are often underspecified since they rely on the very strict criterion that a sup-group can only be assumed if a shared innovation can be demonstrated.

Within the past twenty-five years, the new field of \emph{computational historical linguistics} has emerged which uses methods adapted from computational biology to produce binary-branching phylogenetic trees of languages. The starting point for these methods are parallel word lists, such as the Swadesh list or similar collections of basic vocabulary items. From these lexical data, discrete (usually binary) features are extracted that classify languages in various ways. These features are ideally historically inert, i.e., their values are mostly inherited faithfully from an ancestor language to its descendants. If a mutation occurs, it is passed on to the descendants. If a sufficient number of such features is available, the tree can be automatically reconstructed.

The most widely used type of such features are derived from \textbf{cognate classes}, which are sets of words in different languages that are derived from a common ancestor. For instance, the English word \emph{mother} and the Hindi word \emph{ma:ta:} are cognates, as they both derive from the Proto-Indo-European word \emph{*m\'{e}h$_{\text{2}}$t\={e}r}. All languages sharing a word for the concept of \emph{mother} that is derived from the same Proto-Indo-European word have the value 1 for this feature, while all languages that do not share this cognate have the value 0.

This approach, while widely used, has two limitations. First, it is based on \emph{manual cognacy annotation}, which is a time-consuming and error-prone process. Second, cognate classes are, by definition, confined to a single language family. According to the classical comparative method, a language family is a maximal group of languages for which a common ancestor can be demonstrated, and a cognate class is a set of words for which a common ancestor can be reconstructed. Therefore it is not possible to have cognate classes that span multiple language families. This is a serious limitation, as it means that PCMs cannot be used for comparative studies spanning multiple language families.

There are various proposals in the literature to overcome these limitations. E.g, features can be extracted from the phonetic transcriptions of basic vocabulary word lists directly via machine learning, thereby sidestepping both above-mentioned problems.

In this study, the phylogenetic tree of the world's languages from \citet{jaeger2026worldtree} is used. It is inferred from Lexibank word lists \citep{list2022lexibank} for 105 basic concepts without cognate annotation: the words for each concept are aligned across languages, the alignments are recoded as binary characters, and a maximum-likelihood tree is inferred from these characters and converted to ultrametric form. The tree covers 3,399 languages. Its agreement with the expert classification of Glottolog is good though not perfect: of the 84 Glottolog families with at least four languages in the tree, 59 form a clade. The details of the procedure are given in the original publication.

For both case studies, this tree was pruned to the languages of the sample, i.e., 422 languages for the affix--adposition study and 1,182 languages for the population--inventory study. Languages not covered by the tree were excluded from the samples. Pruning does not change the branch lengths. They are expressed in units of the median branch length of the full tree, so the time scale is the same in both studies and does not depend on the sample.
\begin{figure}
    \centering
    \includegraphics[width=.7\textwidth]{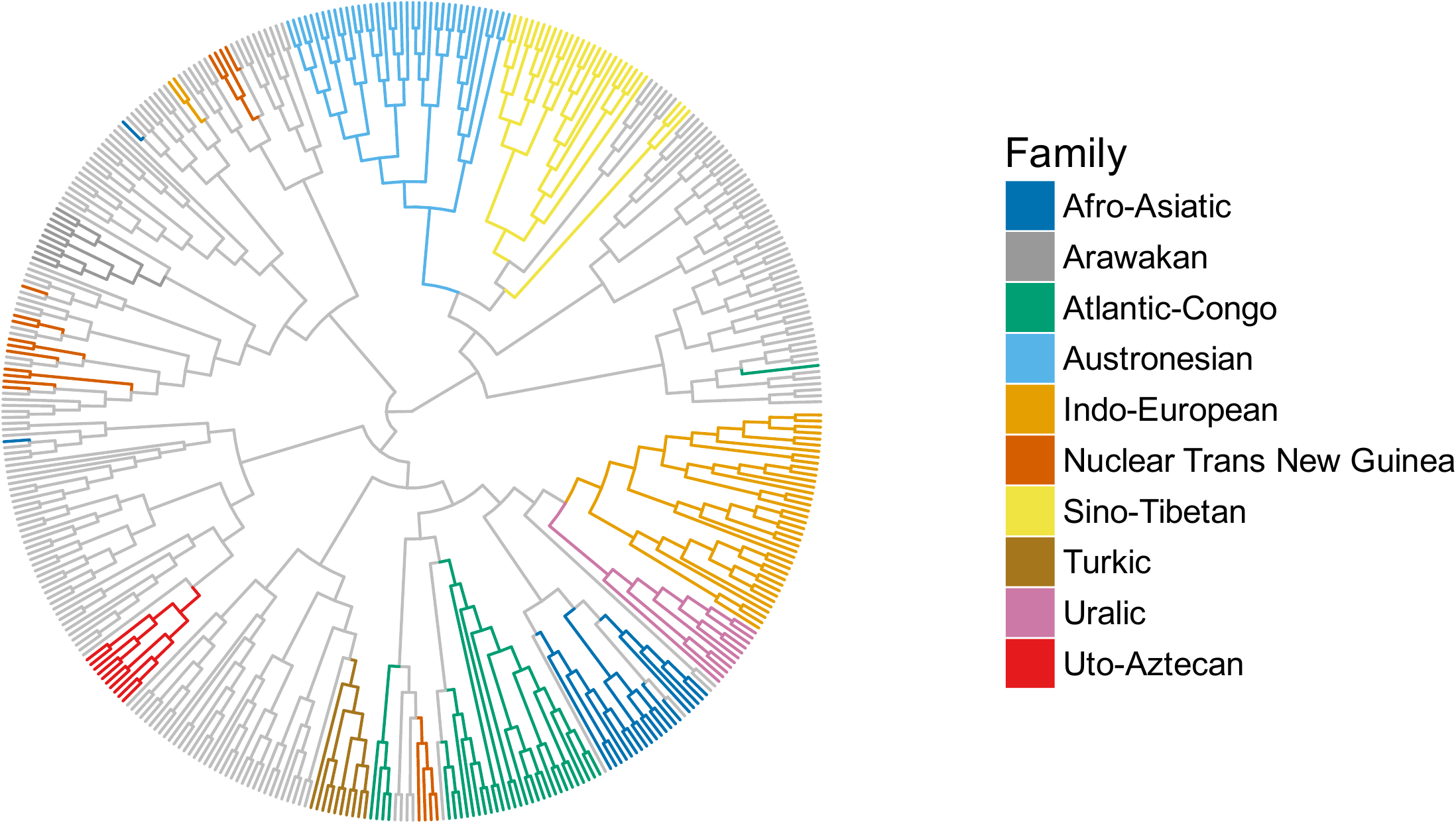}
    \includegraphics[width=.7\textwidth]{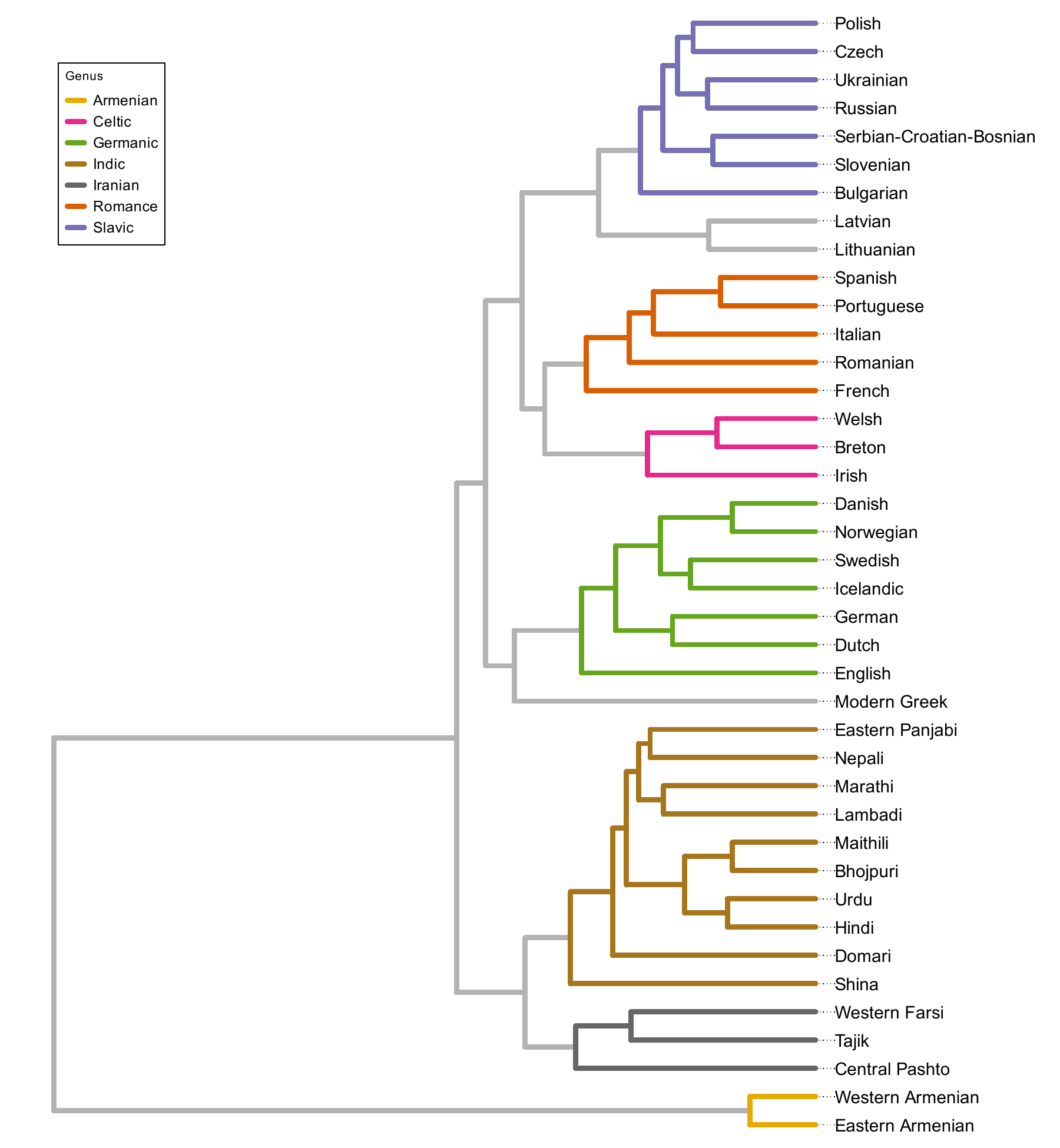}
    \caption{\textbf{Top:} Phylogenetic tree of the world's languages. The tree was pruned to the 422 languages of the affix--adposition study. \textbf{Bottom:} Subtree for the Indo-European languages.}
    \label{fig:world_tree}
\end{figure}

The tree, as well as the sub-tree for the Indo-European language family, is shown in Figure \ref{fig:world_tree} for illustration. It illustrates both strengths and weaknesses of the method. On the one hand, it is clearly visible that the tree captures the main genealogical patterns adequately, largely identifying language families and their internal sub-groupings. For instance, the Austronesian, Uralic, Turkic, Uto-Aztecan and Arawakan languages of the sample each form a clade, and so do the Indo-European languages with the exception of Armenian. On the other hand, many details are at odds with received scholarship. The Afro-Asiatic and the Atlantic-Congo languages are split into seven and five separate groups respectively, and the 18 Nuclear Trans New Guinea languages are spread over ten groups. In the Indo-European sub-tree, the major branches are recovered, but the placement of Greek as the sister of Germanic is arguably historically incorrect, and so is the position of Armenian outside the clade of the other Indo-European languages.

Also, it should be kept in mind that the branch lengths in the tree are not calibrated with information from the historical and archaeological record. They are derived from the amount of lexical change, which was converted into relative time under a model of smoothly varying rates of change. This is correlated with, but not identical to historical time.

In the context of the present study, the purpose of the phylogenetic tree is not to be a faithful representation of the history of language diversification. Rather, it is part of a statistical model, and it serves as a representation of the expected statistical dependencies between languages. In other words, the tree represents the the similarity patterns of core vocabulary items, and it is used as a proxy for the expected similarity patterns of the features under investigation.

There are actually two questions that can be addressed when using phylogenetic control for an analysis of the association between two typological variables:

\begin{itemize}
    \item When assuming that the response variable evolves on the tree, does knowledge of the predictor variable help to predict the response variable?
    \item When assuming that both variables evolve on the tree, is there a correlation between the diachronic changes of the two variables?
\end{itemize}

The first question is addressed by \emph{phylogenetic regression} models, which assume that the response variable evolves on the tree, and that the predictor variable is a fixed effect. The second question is addressed by \emph{phylogenetic correlation} models, which assume that both variables evolve on the tree.

As argued above, for observational data such as the ones dealt with here, a phylogenetic regression model is not appropriate. Rather, the distributions of both variables, and the dependency between them, should be modeled. This can be achieved via a bivariate phylogenetic correlation model.

In this study, it will generally be assumed that continuous variables evolve along the branches of a tree following the \emph{Ornstein-Uhlenbeck process} \citep{uhlenbeck1930theory,hansen1997stabilizing}. 

The Ornstein-Uhlenbeck process (OU process henceforth) is a stochastic process that describes the evolution of a continuous trait over time. The OU process can be described as random walk on a leash. The trait evolves according to a random walk, but it is also subject to a restoring force that pulls it back towards a stable equilibrium value. This means that the trait tends to fluctuate around a mean value, rather than drifting away from the origin indefinitely.

It is characterized by three parameters: the long-term average $\mu$, the \emph{drift} parameter $\lambda$, which describes the tendency of the trait to return to its mean value, and the \emph{diffusion} parameter $\sigma$, which describes the amount of random variation in the trait. When it has value $x_0$ at time $t=0$, the probability density function of the trait at time $t$ is given by:

\begin{align}
    x_t &\sim \mathcal{N} \left( \mu + (x_0 -\mu)e^{-\lambda t}, 
    \frac{\sigma^2}{2\lambda} (1 - e^{-2\lambda t})  \right)
    \label{eq:ou}
\end{align}

The long-term equilibrium distribution when $t \to \infty$ is given by:
\begin{align}
    x_t &\sim \mathcal{N}\left(\mu, \frac{\sigma^2}{2\lambda}\right)
    \label{eq:ou_equilibrium}
\end{align}

The model assumes that the value of the trait (latent variables for the affix-adposition study and log-transformed population size and phoneme inventory size for the population size study) is drawn at random from the equilibrium distribution (\ref{eq:ou_equilibrium}). The value of the trait at a daughter node follows the distribution in (\ref{eq:ou}) with the value at the parent node as $x_0$ and the length of the branch as $t$. The parameters $\mu$, $\lambda$, and $\sigma$ are estimated from the data.

For a given phylogenetic tree, the predicted covariance of a trait value evolving according to OU between two languages $i$ and $j$ is given by:

\begin{align}
    \text{Cov}(x_i, x_j) &= \frac{\sigma^2}{2\lambda} e^{-\lambda t_{ij}},
\end{align}

where $t_{ij}$ is the length of the path between languages $i$ and $j$ in the tree. The resulting variance-covariance matrix denoted by $\Sigma_{OU}$.

Since the phylogenetic tree used here is possibly unreliable above the family level, this phylogenetic component is combined with family-level intercepts. The model specification is given in Model \ref{mod:logistic_ou}.

\begin{model}[htbp]
    \centering
    \begin{minipage}{.9\linewidth}
\begin{align}
    \rho_f, \rho_l &\sim \text{Uniform}(-1, 1) \\
    \sigma_{f,1}, \sigma_{f,2}, \sigma_{l,1}, \sigma_{l,2}, \lambda_1, \lambda_2&\sim \text{LogNormal}(0, 1) \\
    \Sigma_f &= \begin{pmatrix}
        \sigma_{f,1}^2 & \rho_f \sigma_{f,1} \sigma_{f,2} \\
        \rho_f \sigma_{f,1} \sigma_{f,2} & \sigma_{f,2}^2
    \end{pmatrix} \\
    \mu_{\text{adp}} &\sim \mathcal{N}(0, 2), \quad \mu = (0, \mu_{\text{adp}}) \\
    z_{f,j} &\sim \mathcal{N}(\mathbf{0}, \Sigma_f) \quad \text{for each family } j \\
    w_{\text{root}} &\sim \mathcal{N}(\mu, S(\infty)) \\
    w_{c} \mid w_{p} &\sim \mathcal{N}\left(\mu + e^{-\Lambda t_c}(w_{p} - \mu),\ S(t_c)\right) \quad \text{for each branch } p \to c \\
    S(t)_{kk'} &= \rho_{kk'}\,\sigma_{l,k}\,\sigma_{l,k'}\,\frac{1 - e^{-(\lambda_k + \lambda_{k'})t}}{\lambda_k + \lambda_{k'}}, \quad \rho_{11} = \rho_{22} = 1,\ \rho_{12} = \rho_{21} = \rho_l \\
    z_i & := z_{f,f_i} + w_{i} \\
    c_1 &\sim \mathcal{N}(0, 2) \\
    \zeta_k &\sim \text{LogNormal}(0, 2), \quad c_{k+1} = c_k + \zeta_k + 10^{-2} \\
    F_{i,k} &= \frac{1}{1 + e^{-(c_k - z_{\text{affix},i})}} \quad \text{for } k = 1,\ldots,4 \\
    \text{affix}_i &\sim \text{Categorical}(F_{i,1},\ F_{i,2} - F_{i,1},\ F_{i,3} - F_{i,2},\ F_{i,4} - F_{i,3},\ 1 - F_{i,4}) \\
    \text{preposition}_i &\sim \text{Bernoulli}\left(\frac{1}{1 + e^{-z_{\text{adposition},i}}}\right)
\end{align}
\end{minipage}
\caption{Phylogenetic correlation model for affix–adposition association. $w$ is the pair of latent traits at the nodes of the tree, $t_c$ the length of the branch leading to node $c$, $\Lambda = \mathrm{diag}(\lambda_1, \lambda_2)$, and $w_i$ the value at the tip for language $i$.}
\label{mod:logistic_ou}
\end{model}

The two latent traits evolve jointly along the branches of the tree. Each trait has its own mean-reversion rate $\lambda_k$ and diffusion rate $\sigma_{l,k}$, and $\rho_l$ is the correlation between the random changes of the two traits. The transition covariance $S(t)$ is the exact covariance of a bivariate OU process over a branch of length $t$, and $S(\infty)$ is its stationary covariance. The long-term mean of the affix latent is fixed at 0, since the cutpoints already determine the overall distribution of affix types. The long-term mean $\mu_{\text{adp}}$ of the adposition latent plays the role of the intercept $\alpha$ in the previous models.

The model was fitted using the \emph{rstan} package in R. The results are summarized in Table~\ref{tab:ou_bivariate_model}. As in the hierarchical model, one finds clear evidence for a positive correlation between affix type and adposition type at the family level (mean $\rho_f$ = 0.726, $95\%$ credible interval $(0.463, 0.919)$). Unlike in the hierarchical model, the correlation at the language level is also credibly positive (mean $\rho_l$ = 0.349, $95\%$ credible interval $(0.065, 0.622)$).

Again, this model constitutes a clear improvement over the previous models, with a log-Bayes factor of approximately 27 in favor of the phylogenetic model in comparison to the hierarchical model.

\begin{table}[h!]
    \centering
    \begin{tabular}{@{}lrrrrrrr@{}}
        \toprule
        \textbf{Parameter} & \textbf{Mean} & \textbf{SE\_Mean} & \textbf{SD} & \textbf{2.5\%} & \textbf{97.5\%} & \textbf{n\_eff} & \textbf{Rhat} \\
        \midrule
        cutpoints$_1$ & -0.192 & 0.006 & 1.216 & -2.722 & 2.226 & 37571 & 1.00 \\
        cutpoints$_2$ & 2.570 & 0.026 & 1.646 & -0.148 & 6.330 & 3864 & 1.00 \\
        cutpoints$_3$ & 6.672 & 0.064 & 2.960 & 2.802 & 13.797 & 2150 & 1.00 \\
        cutpoints$_4$ & 10.290 & 0.097 & 4.269 & 5.033 & 20.706 & 1947 & 1.00 \\
        $\mu_{\text{adp}}$ & -1.076 & 0.010 & 1.909 & -4.699 & 2.736 & 35442 & 1.00 \\
        $\sigma_{l,1}$ & 2.552 & 0.028 & 1.239 & 1.035 & 5.579 & 1936 & 1.00 \\
        $\sigma_{l,2}$ & 2.633 & 0.023 & 1.170 & 1.172 & 5.694 & 2686 & 1.00 \\
        $\lambda_1$ & 0.135 & 0.000 & 0.047 & 0.056 & 0.237 & 24894 & 1.00 \\
        $\lambda_2$ & 0.051 & 0.000 & 0.022 & 0.018 & 0.102 & 33362 & 1.00 \\
        $\sigma_{f,1}$ & 5.500 & 0.054 & 2.259 & 2.813 & 11.004 & 1761 & 1.00 \\
        $\sigma_{f,2}$ & 9.559 & 0.095 & 4.224 & 4.391 & 20.586 & 1981 & 1.00 \\
        $\rho_l$ & 0.349 & 0.001 & 0.142 & \textbf{0.065} & \textbf{0.622} & 22515 & 1.00 \\
        $\rho_f$ & 0.726 & 0.001 & 0.117 & \textbf{0.463} & \textbf{0.919} & 13399 & 1.00 \\
        \bottomrule
    \end{tabular}
    \vspace{1em}
    \caption{Posterior summaries for the bivariate Ornstein--Uhlenbeck model with family-level correlation \(\rho_f\), correlation \(\rho_l\) between the random changes of the two latent traits, and mean-reversion rates \(\lambda_1, \lambda_2\). Cutpoints map the latent variable for the ordinal response to observed categories.}
    \label{tab:ou_bivariate_model}
\end{table}

It is important to appreciate that the language-level correlation holds between the diachronic changes of the two variables. Informally put, if a language evolves towards a more prefixing type for inflectional morphology, it tends to become more likely to be prepositional, and vice versa.

\begin{figure}
    \centering
    \includegraphics[width=\textwidth]{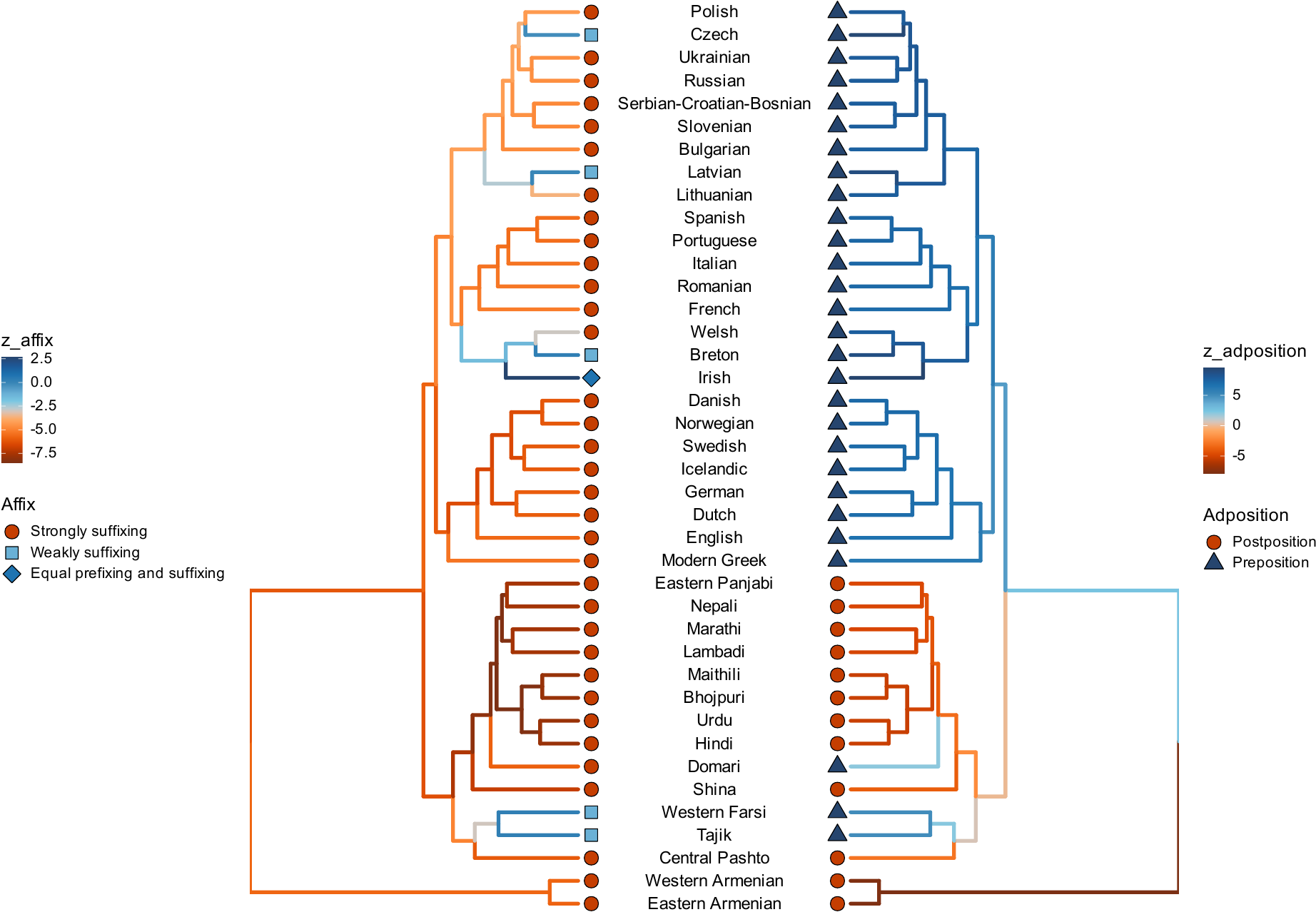}
    \caption{Ancestral state reconstructions for the Indo-European sub-tree. The left panel shows the reconstructed values for the affix latent variable, while the right panel shows the reconstructed values for the adposition latent variable. Markers at the leaves indicate observed features.}
    \label{fig:ancestral_states}
\end{figure}

It is an advantage of an explicit phylogenetic model that ancestral state reconstruction is automatically conducted as a side effect. These reconstructions can be extracted from the fitted model. Figure \ref{fig:ancestral_states} shows the reconstructed ancestral states for both latent variables for the Indo-European sub-tree for illustration.

The tendency for correlated evolution can be observed, for instance, within the Indo-Iranian and the Celtic subfamilies.

The fact that the correlation is positive both between and within families is consistent with the uniformitarian hypothesis, according to which the same mechanisms driving evolution within families are also responsible for the distribution across families. Therefore one can conclude with high confidence that a preference for prepositions is universally associated with a preference for prefixes, and vice versa.

In the second case study, the model setup is similar, except that the co-evolving continuous variables are directly observed here. Since there are no cutpoints, the long-term means of both variables are free parameters. The model specification is given in Model \ref{mod:soundpop_ou}.

\begin{model}[htbp]
    \centering
    \begin{minipage}{.9\linewidth}
        \begin{align}
            \rho_f, \rho_l &\sim \text{Uniform}(-1, 1) \\
            \sigma_{f,1}, \sigma_{f,2}, \sigma_{l,1}, \sigma_{l,2}, \lambda_1, \lambda_2 &\sim \text{LogNormal}(0, 1) \\
            \Sigma_f &= \begin{pmatrix}
                \sigma_{f,1}^2 & \rho_f \sigma_{f,1} \sigma_{f,2} \\
                \rho_f \sigma_{f,1} \sigma_{f,2} & \sigma_{f,2}^2
            \end{pmatrix} \\
            \mu &\sim \mathcal{N}(\mathbf{0}, 4\mathbf{I}) \\
            z_{f,j} &\sim \mathcal{N}(\mathbf{0}, \Sigma_f) \quad \text{for each family } j \\
            w_{\text{root}} &\sim \mathcal{N}(\mu, S(\infty)) \\
            w_{c} \mid w_{p} &\sim \mathcal{N}\left(\mu + e^{-\Lambda t_c}(w_{p} - \mu),\ S(t_c)\right) \quad \text{for each branch } p \to c \\
            S(t)_{kk'} &= \rho_{kk'}\,\sigma_{l,k}\,\sigma_{l,k'}\,\frac{1 - e^{-(\lambda_k + \lambda_{k'})t}}{\lambda_k + \lambda_{k'}}, \quad \rho_{11} = \rho_{22} = 1,\ \rho_{12} = \rho_{21} = \rho_l \\
            x_i &:= w_i + z_{f,f_i}
        \end{align}
    \end{minipage}
    \caption{Phylogenetic correlation model for segment inventory/population size association. $x_i$ is the pair of standardized log population size and log inventory size of language $i$; $w$, $t_c$ and $\Lambda$ are as in Model~\ref{mod:logistic_ou}.}
    \label{mod:soundpop_ou}
\end{model}

As in Model~\ref{mod:logistic_ou}, each variable has its own mean-reversion rate. The covariance between related languages therefore decays at a different rate for the two variables, and $\rho_l$ is the correlation between the random changes of population size and inventory size along the branches of the tree.

The model was fitted using the \emph{rstan} package in R. The results are summarized in Table~\ref{tab:bivariate_phylo_model}. As in the hierarchical model, one finds clear evidence for a positive correlation between population size and phoneme inventory size at the family level (mean $0.58$, credible interval $(0.34, 0.78)$), but no evidence for a correlation between the diachronic changes of the two variables along the branches of the tree: $\rho_l$ has a posterior mean of $-0.04$, and its credible interval $(-0.103, 0.021)$ includes zero. 

Statistical model comparison clearly favors the phylogenetic model, with a log-Bayes factor of approximately 133 in comparison to the hierarchical model.

\begin{table}[h!]
    \centering
    \begin{tabular}{@{}lrrrrrrr@{}}
        \toprule
        \textbf{Parameter} & \textbf{Mean} & \textbf{SE\_Mean} & \textbf{SD} & \textbf{2.5\%} & \textbf{97.5\%} & \textbf{n\_eff} & \textbf{Rhat} \\
        \midrule
        $\sigma_{l,1}$ & 0.483 & 0.000 & 0.020 & 0.446 & 0.524 & 2605 & 1.00 \\
        $\sigma_{l,2}$ & 0.494 & 0.001 & 0.022 & 0.453 & 0.540 & 1929 & 1.00 \\
        $\lambda_1$ & 0.215 & 0.001 & 0.029 & 0.160 & 0.275 & 3084 & 1.00 \\
        $\lambda_2$ & 0.226 & 0.001 & 0.033 & 0.165 & 0.297 & 2398 & 1.00 \\
        $\mu_1$ & -0.474 & 0.001 & 0.118 & -0.708 & -0.238 & 13542 & 1.00 \\
        $\mu_2$ & -0.327 & 0.001 & 0.118 & -0.563 & -0.095 & 14491 & 1.00 \\
        $\sigma_{f,1}$ & 0.568 & 0.001 & 0.062 & 0.448 & 0.693 & 5026 & 1.00 \\
        $\sigma_{f,2}$ & 0.759 & 0.001 & 0.068 & 0.628 & 0.898 & 9366 & 1.00 \\
        $\rho_l$ & -0.041 & 0.000 & 0.032 & \textbf{-0.103} & \textbf{0.021} & 8753 & 1.00 \\
        $\rho_f$ & 0.575 & 0.002 & 0.113 & \textbf{0.339} & \textbf{0.777} & 3877 & 1.00 \\
        \bottomrule
    \end{tabular}
    \vspace{1em}
    \caption{Posterior summaries for parameters in the bivariate phylogenetic OU model with family-level correlation. The 95\% credible interval of the correlation $\rho_l$ between the random changes of the two variables includes 0.}
    \label{tab:bivariate_phylo_model}
\end{table}

This result is in stark contrast to the findings of the affix-adposition association. No evidence for a correlated diachronic evolution of population size and segment inventory size is found. But how to account for the family-level correlation?

When two variables are correlated, two causal scenarios have to be considered. The correlation may be due to a direct causal relationship between the two variables, or it may be due to a third variable exerting causal influence on both observed variables. The absence of small-scale coevolution essentially excludes the possibility of a direct causal link. Therefore, the family-level correlation must be due to a third variable. This common cause is arguably connected to geography, paired with contingent historical events.

These findings reinforce the conclusions reached by \cite{moranetal12} via a hierarchical regression model.

\subsection{Summary}

The results of the two case studies illustrate the importance of controlling for genealogical dependencies in typological data. The association between affix position and adposition type is a robust diachronic tendency, observable both within and across language families. In contrast, the apparent correlation between phoneme inventory size and population size disappears at the level of individual languages when phylogenetic relationships are taken into account, suggesting that the observed association is a by-product of shared ancestry or geography rather than a result of co-evolution.

It is also instructive to look at the results of statistical model comparison for both studies. Two methods were used: PSIS-LOO (\emph{Pareto-smoothed importance sampling leave-one-out cross-validation}, \citealt{vehtarietal2017}, as implemented in the R-package \emph{loo}, \citealt{loo}) and log-Bayes factors. The results are summarized in Table~\ref{tab:model_comparison_synopsis}. (PSIS-LOO compares the \emph{expected log pointwise predictive density}, elpd; a higher value indicates a better fit to the data.)

Both methods require some care with the models used here, since they contain a large number of latent variables: a pair of latent traits for each language in the affix--adposition models, and family effects and ancestral states in the hierarchical and phylogenetic models. For PSIS-LOO, the latent variables that are informed only by the language left out were integrated out of its likelihood; otherwise leaving the language out removes all information about these variables, and the importance-sampling approximation breaks down. The predictive density thus refers to a new language, not to a new observation of a language whose latent traits are known. For the few languages for which the approximation was nevertheless unreliable, the leave-one-out predictive density was computed exactly: the model was refitted without the data of that language, which otherwise remains in the tree and in its family, and the predictive density of its data under the refitted model replaced the approximation. Log marginal likelihoods were estimated by stepping-stone sampling \citep{xieetal2011,fanetal2011}, and the estimates were cross-checked with thermodynamic integration and, where feasible, with bridge sampling \citep{bridgesampling}. The numerical details (the integration of the latent variables, the sampling schedule, the convergence diagnostics and the results of the cross-checks) are documented in the code repository (see the statement on data and code availability at the end of this chapter).

\begin{table}[ht]
    \centering
    \renewcommand{\arraystretch}{1.3}
    \resizebox{\textwidth}{!}{%
    \begin{tabular}{@{}l
        S[round-mode=places, round-precision=0, table-format=4.0]@{\hspace{1em}}
        S[round-mode=places, round-precision=0, table-format=4.0]
        @{\hspace{2em}}
        S[round-mode=places, round-precision=0, table-format=4.0]@{\hspace{1em}}
        S[round-mode=places, round-precision=0, table-format=4.0]@{}}
        \toprule
        \textbf{Model Type} 
        & \multicolumn{2}{c}{\textbf{Affix–Adposition}} 
        & \multicolumn{2}{c}{\textbf{Segment Inventory–Population Size}} \\
        \cmidrule(lr){2-3} \cmidrule(lr){4-5}
        & \multicolumn{1}{c}{\textbf{Bayes factor (log)}} & \multicolumn{1}{c}{$\Delta$\textbf{elpd}}
 
        & \multicolumn{1}{c}{\textbf{Bayes factor (log)}} & \multicolumn{1}{c}{$\Delta$\textbf{elpd}}
        \\
        \midrule
        \textit{Vanilla Model}      
        & -196 & -273 & -752 & -933 \\
        \textit{Hierarchical Model}
        & -27  & -59  & -133  & -207  \\
        \addlinespace
        \textit{Phylogenetic Model} 
        & 0 & 0 & 0 & 0 \\
        \bottomrule
    \end{tabular}
    }
    \vspace{1em}
    \caption{Model comparison for both case studies. Columns 2–3 correspond to the affix–adposition case study, and columns 4–5 to the segment inventory–population size case study. Higher elpd and log-Bayes values indicate better model fit. All values are differences to the phylogenetic model. The standard errors of the elpd differences are 16.4 (vanilla) and 7.9 (hierarchical) for the affix--adposition models, and 38.6 and 24.7 for the segment inventory--population size models; the Monte Carlo standard errors of the log marginal likelihoods are below 0.3.}
    \label{tab:model_comparison_synopsis}
\end{table}

According to both methods, the fit of the data massively improves when adding family-level random intercepts. Adding phylogenetic control improves the fit further, by a smaller but still decisive margin. This indicates that the added complexity of using the phylogenetic comparative method is clearly justified when conducting typological studies.

\section{Conclusion}

This chapter has reviewed recent advances in computational typology, with a particular focus on the role of quantitative and phylogenetic methods in the investigation of language universals and structural correlations. Classic typological hypotheses, such as Greenberg’s Universal 27, as well as more recent proposals concerning the interaction of linguistic and non-linguistic variables, have been re-examined using statistical models that account for genealogical and areal dependencies.

A central methodological contribution is the application of bivariate models for mixed data types -- including ordinal and binary variables -- embedded in hierarchical and phylogenetic frameworks. These models allow for the joint modeling of two traits, taking into account both family-level structure and phylogenetic inertia via the Ornstein–Uhlenbeck process. This approach facilitates the distinction between correlations driven by shared descent and those that reflect co-evolution under common functional pressures.

The empirical findings illustrate the utility of this approach. The association between affix position and adposition type proves to be a robust diachronic tendency, observable both within and across language families. In contrast, the apparent correlation between phoneme inventory size and population size disappears at the level of individual languages when phylogenetic relationships are taken into account, suggesting that the observed association is a by-product of shared ancestry or geography rather than a result of co-evolution.

Overall, these results highlight the importance of controlling for non-independence in typological data. Computational models that integrate genealogical information offer a principled way to test claims about universals, revealing whether they reflect true evolutionary tendencies or are artifacts of sampling and historical contingency. As typological datasets continue to grow in coverage and detail, such models will play a central role in refining the understanding of cross-linguistic patterns and their underlying causes.

As briefly alluded to above, phylogenetic control does not fully eliminate the problem of non-independence. For instance, it does not take language contact into account. While active research in this direction is underway, the integration of phylogenetic and geostatistical methods is still in its infancy and provides ample opportunities for future research.

\section*{Acknowledgements}

This research was supported by the DFG Centre for Advanced Studies in the Humanities Words, Bones, Genes, Tools (DFG-KFG 2237) and by the European Research Council (ERC) under the European Union's Horizon 2020 research and innovation programme (Grant agreement 834050).

Anthropic's Claude Code was used for proofreading, code review, and the re-implementation of the likelihood computations used for model comparison.

\section*{Data and code availability}

The data and code for the analyses presented in this chapter are available at \url{https://codeberg.org/profgerhard/computational_typology_routledge}. 

\bibliographystyle{apacite}

\end{document}